\pdfoutput=1

\documentclass[final]{article}

\PassOptionsToPackage{numbers, compress}{natbib}

\usepackage{nips_2018}

\usepackage[utf8]{inputenc} 
\usepackage[T1]{fontenc}    
\usepackage{hyperref}       
\usepackage{url}            
\usepackage{booktabs}       
\usepackage{amsfonts}       
\usepackage{nicefrac}       
\usepackage{microtype}      

\newcommand{\Loss}{\mathcal{L}}

\usepackage{amsmath}
\usepackage{subfig}
\usepackage{multirow}
\usepackage{graphicx}

\DeclareMathOperator*{\argmin}{arg\,min}

\title{Cluster-Based Learning from Weakly Labeled Bags in Digital Pathology}

\author{
  Shazia Akbar, Anne L. Martel \\
  Sunnybrook Research Institute \\
  Medical Biophysics, University of Toronto\\
  Vector Institute, Toronto, Canada \\
  \texttt{\{shazia.akbar,anne.martel\}.sunnybrook.ca} \\
}

\begin{document}

\maketitle

\begin{abstract}
 To alleviate the burden of gathering detailed expert annotations when training deep neural networks, we propose a weakly supervised learning approach to recognize metastases in microscopic images of breast lymph nodes. We describe an alternative training loss which clusters weakly labeled bags in latent space to inform relevance of patch-instances during training of a convolutional neural network. We evaluate our method on the Camelyon dataset which contains high-resolution digital slides of breast lymph nodes, where labels are provided at the image-level and only subsets of patches are made available during training.   
\end{abstract}

\section{Introduction}

With advancements in imaging and hardware, machine learning techniques have been well adopted in the medical domain for analysis of highly complex data, however one of the remaining bottlenecks is the task of gathering large volumes of annotated training data.
This is largely due to costs of expert annotators and time constraints associated with fine-tooth combing of large datasets. Specifically in medicine, expert annotators require several years of training and are currently bombarded with the large volume of healthcare data generated on a routine basis. Furthermore, due to intra- and inter-rater variability, training instances may need to be annotated by multiple observers afterwhich there is uncertainty surrounding the usage of multiple ``ground truth'' labels. With the growing success of deep neural networks, there is now a higher demand for large annotated datasets in the medical domain to prevent overfitting. To overcome this problem, we propose a method of using weak labels in a deep learning framework. The goals of our approach is two-fold: 1) identify class labels at the instance-level given labels provided at a much coarser level, and 2) learn features in a deep network with limited expert input. 

Here, we tackle the problem of identifying tumor metastases in microscopic images in the publicly available Camelyon challenge dataset \cite{Camelyon2016}. In our setup, training instances are patches extracted from images with extremely high dimensions, 200,000 x 100,000 pixels, and weak labels are assigned at the image-level. In reported experiments, this equates to over 80,000 training patch-instances per label, with $<5\%$ of these correlating to true positive (i.e. tumor) patch-instances. As such, the class imbalance is extremely high and this trend continues to other modalities such as magnetic resonance imaging \cite{Dubey2014} and experimental methods for cancer diagnosis and prognosis such as miRNA \cite{Kothandan2015}.

In this paper, we take inspiration from multiple instance learning (MIL) \cite{Dietterich1997} to outline an approach for training a convolutional neural network (CNN) with image-level annotations. In the first phase of our pipeline, features are learned in an unsupervised manner via a variational autoencoder, which is subsequently used to identify clusters of patch-instances in feature space. In the case of microscopic images, such features may correlate to stroma, fatty tissue or lumen. We then identify ``cluster-classes'' in a latent representation to distinguish between patch-instances without labels, and adjust the training loss appropriately. We show that when weak labels are adopted in a deep learning framework, our proposed method, CCE$^{+}$, can achieve performance comparable to a fully supervised approach and, furthermore, maintains performance when distracting training samples are introduced during training.


\section{Method}
\label{sec:method}

In the traditional MIL framework, ground truth consists of labels provided at the \emph{bag-level}, where each bag contains $N$ training instances, where $N>1$. Bags are then labeled as either positive, in which at least one instance within the bag must be positive, or negative in which all instances in the bag are negative. Here, we use a similar framework where large histology images (hereon refered to as digital slides) denote bags, and patches extracted from each slide as patch-instances. Digital slides which contain metastases are labeled as positive, and healthy digital slides as negative. The proposed framework is not limited to binary classification and can also be used in a multi-class schema.

In a typical MIL pipeline, training instances are represented as feature vectors of the original data source which are cross-examined during training. The objective is to estimate a classification function that provides the likelihood that an instance in a positive bag has a true positive label \cite{Amores2013}. However one of key limitations of using MIL in a deep learning framework is that all training instances within each bag cannot be made available during training. Due to memory contraints, when training a deep model, instances are made available in batches and therefore only a subset of each ``bag'' is made available in a single batch, eliminating the possibility of performing cross-correlations within a bag as is done in previous work \cite{Sun2016}. We combat this by using unsupervised learned features to inform the model of patch-instances which share similarities based on appearance and texture, regardless of the grade or progression of tumor present in the image. The model therefore has two types of input when computing the loss of training instances: 1) predictions generated from the model in its current state, and 2) a latent representation of each patch-instance.

A common loss function used in classification tasks is the categorical cross-entropy (CCE) which combines predictions from the output of a network ($p$), typically a softmax layer, with ground truth labels ($y$):
\begin{equation}
\text{CCE}(y,p)=-\sum\limits_{i} y_i log(p_i) 
\end{equation}

However when labels are noisy, CCE alone is uninformative and inhibits learning. Here we propose learning an estimated ground truth from an unsupervised representation, which we term a cluster-class ($\hat{y}$). We opt to train a variational autoencoder (VAE) in which a Gaussian latent representation is learned from image data alone, however any other unsupervised learning technique can be substituted.


To estimate cluster-classes we perform $K$-means clustering in the latent space of our trained VAE. From the weak bag-labels in each cluster, we perform a majority vote to compute a cluster-class label, denoted as $C^i, i=\{1...K\}$. During training, the cluster label of each patch-instance is denoted as the nearest cluster in feature space, giving us an estimated cluster-class, $\hat{y}=C^{\argmin{D(f, f_j)}}$, where $f$ is the feature encoding of a given patch and $f_j$ are the cluster centers. Our final loss is then a measure of both traditional CCE, and the CCE of predicted outcomes and cluster-classes, weighted by $\alpha$, set to $0.5$ in reported experiments. 

\begin{equation}
\Loss = \alpha \text{CCE}(y, p) + (1-\alpha) \text{CCE}(\hat{y}, p)
\end{equation}



\section{Experiments and Results}

\subsection{MNIST-BAG}

\begin{figure}
\centering
    \begin{tabular}{c|c|c|c|c}
        {\bf 1} & {\bf 3} & {\bf 8} & {\bf 2} & {\bf 0} \\
        & & & & \\
	    \includegraphics[width=0.15\textwidth]{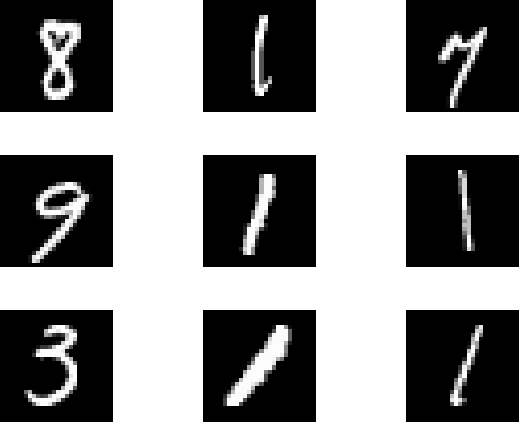} &
	    \includegraphics[width=0.15\textwidth]{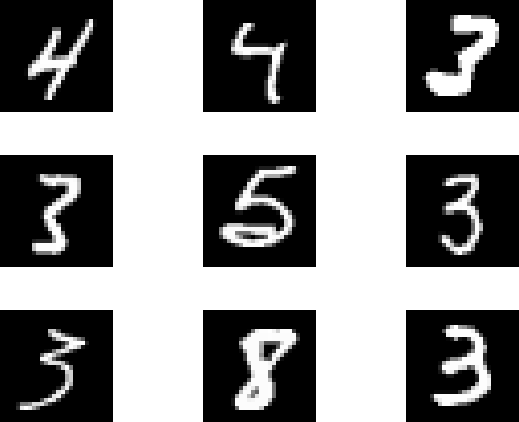} &
	    \includegraphics[width=0.15\textwidth]{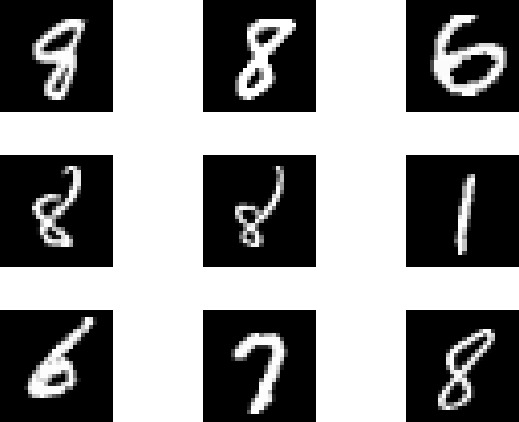} &
	    \includegraphics[width=0.15\textwidth]{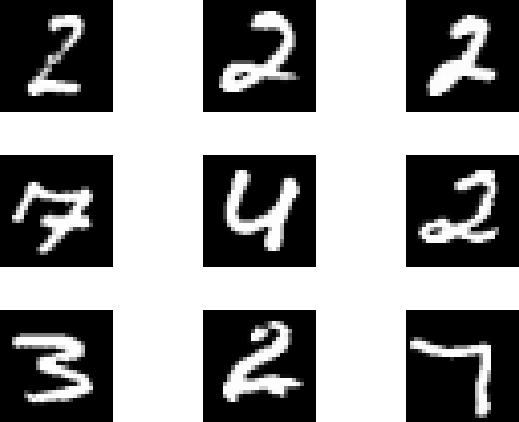} & 
	    \includegraphics[width=0.15\textwidth]{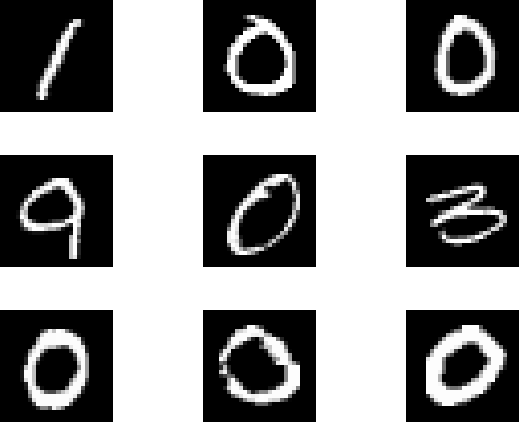} 
	\end{tabular}
	\caption{Subset of MNIST-BAG dataset ($N=9$) with corresponding bag labels (top row).}
	\label{fig:mnistbag}
\end{figure}

As a toy example, we created a MNIST-BAG dataset (Figure \ref{fig:mnistbag}) containing bags of MNIST digits where at least half the instances within each bag were reflective of the same label and the remaining bag contained distracting MNIST digits i.e. digits not equal to the bag label. MNIST-BAG also demonstrates how our method can be applied to a multi-class problem set.

Our experimental setup is shown in Table \ref{tab:experiments}. We compared traditional categorical cross entropy (CCE) loss to our proposed method (CCE$^{+}$) and also varied the number of clusters ($N$) learned from the latent representation (Table \ref{tab:mnist}). As expected, as the number of instances in each bag increases, accuracy rates using CCE fell with a significant drop at $N=200$. CCE$^{+}$ was able to maintain performance regardless of the number of instances in each bag.

\begin{table}
\begin{center}
\subfloat{
     \centering
		\begin{tabular}{l|c|c|c|}
			\multicolumn{1}{c}{} & \multicolumn{1}{c}{MNIST} & \multicolumn{1}{c}{} & \multicolumn{1}{c}{Camelyon} \\ \cline{2-2} \cline{4-4}
			{\bf VAE:} &  {32 3x3 conv} & {} & {64 3x3 conv} \\ \cline{2-2} \cline{4-4}
			{} &  {maxpool} & {} & {maxpool} \\ \cline{2-2} \cline{4-4}
			{} &  {32 3x3 conv} & {} & {128 3x3 conv} \\ \cline{2-2} \cline{4-4}
			{} &  {maxpool} & {} & {maxpool} \\ \cline{2-2} \cline{4-4}
			{} &  {512 dense} & {} & {64 3x3 conv} \\ \cline{2-2} \cline{4-4}
			{} &  {64 encoding} & {} & {maxpool} \\ \cline{4-4}
			{} &  {...} & {} & {64 3x3 conv} \\ \cline{2-2} \cline{4-4}
			{} &  {28 x 28 x 1} & {} & {256 dense} \\ \cline{2-2} \cline{4-4}
			\multicolumn{1}{c}{} &  \multicolumn{1}{c}{} & {} & {24 encoding} \\ 
			\multicolumn{1}{c}{} &  \multicolumn{1}{c}{} & {} & {...} \\ \cline{4-4}
			\multicolumn{1}{c}{} &  \multicolumn{1}{c}{} & {} & {256 x 256 x 3} \\ \cline{4-4}
		\end{tabular}
}
\subfloat{
    \centering
	\begin{tabular}{l|c|c|c|}
		\multicolumn{1}{c}{} & \multicolumn{1}{c}{MNIST} & \multicolumn{1}{c}{} & \multicolumn{1}{c}{Camelyon} \\ \cline{2-2} \cline{4-4}

			{\bf CNN:} &  {64 3x3 conv} & {} & {InceptionNet \cite{Szegedy2016}} \\ \cline{2-2} 
		{} &  {maxpool} & {} & {...} \\ \cline{2-2} \cline{4-4}
		{} &  {32 3x3 conv} & {} & {global avg. pool} \\ \cline{2-2} \cline{4-4}
		{} &  {maxpool} & {} & {dropout ($p=0.5$)} \\ \cline{2-2} \cline{4-4}
		{} &  {dropout ($p=0.2$)} & {} & {softmax (2)} \\ \cline{2-2} \cline{4-4}
		{} &  {128 dense} & \multicolumn{1}{c}{} & \multicolumn{1}{c}{} \\ \cline{2-2} 
		{} &  {softmax (10)} & \multicolumn{1}{c}{} & \multicolumn{1}{c}{} \\ \cline{2-2} 
		\multicolumn{4}{c}{} \\
		\multicolumn{4}{c}{} \\
		\multicolumn{4}{c}{} \\
		\multicolumn{4}{c}{} \\
	\end{tabular}
}
\end{center}
\caption{Network architectures of VAEs and CNNs for MNIST-BAG and Camelyon datasets. The last layer of the VAE is the same dimension as the input patch-instances.}
\label{tab:experiments}
\end{table}

\begin{table}
\begin{center}
\begin{tabular}{|l|ccccc|}
\hline
\multirow{2}{*}{Method} & \multicolumn{5}{c|}{Number of instances per bag ($N$)} \\ \cline{2-6}
{} & {\bf 10} & {\bf 50} & {\bf 100} & {\bf 200} & {\bf 500} \\ \hline
{CCE} & {0.9777} & {0.9409} & {0.9364} & {0.5605} & {0.6577} \\
{CCE$^{+}$, $K=10$} & {0.7938} & {0.7860} & {0.7777} & {0.7532} & {0.7525} \\
{CCE$^{+}$, $K=20$} & {0.9222} & {0.9233} & {0.9231} & {0.8524} & {0.8398} \\
{CCE$^{+}$, $K=100$} & {\bf 0.9791} & {\bf 0.9712} & {\bf 0.9425} & {\bf 0.9368} & {\bf 0.9395} \\
\hline
\end{tabular}
\end{center}
\caption{Accuracy rates on MNIST-BAG using two loss functions: categorical cross-entropy (CCE) and our proposed cluster-based CCE (CCE$^{+}$).}
\label{tab:mnist}
\end{table}

Also, as the number of clusters were increased, so did test accuracy performance in MNIST-BAG. Saying this, at $N=200$ and with only $10$ clusters, CCE$^{+}$ surpassed traditional CCE. With $100$ clusters, test accuracy rates were higher than CCE regardless of the size of each bag, suggesting we can use CCE$^{+}$ without sacrificing on quality if sufficient number of clusters are extracted from the latent representation. When $\alpha=0$ i.e. when classification was performed based on VAE features alone, we achieve an accuracy rate of $0.9565$ with $200$ clusters, slightly lower than CCE alone ($0.9791$), suggesting information from both the bag-level label and the VAE latent space is most beneficial.

\subsection{Camelyon}


We used the Camelyon 2016 challenge dataset \cite{Camelyon2016} to evaluate our proposed method which is composed of 400 whole slide images (WSIs) of sentinel lymph node acquired from two different sites. $270$ ($160$ normal and $110$ containing
metastases) of these slides make up a fully annotated training set, however here we only used the image-level labels to train our network. Due to the pyramidal data format of histology slides, we use OpenSlide \cite{Goode2013} to read digital slides at x10 objective.
To validate tumor localization we used WSIs which contained metastases in the independent test set, totalling $49$ WSIs. 

{\bf Experimental Setup:}
We created bags by extracting 256x256 RGB patches from WSIs and labeled them with their corresponding WSI label: $1$ for WSIs which contained metastases and $0$ otherwise. $100$ patches were randomly extracted from each WSI within regions containing tissue. The background was eliminated by thresholding the hue color channel and eliminating spurious regions. This method was derived from \cite{Valkonen2017} and variations on this method has been widely adopted in digital pathology.
Patches were randomly selected in each epoch to sufficiently sample from each WSI over the duration of training. During testing, only patches within tissue regions were evaluated.
Each bag of $100$ training instances was also balanced to contain roughly equal numbers of cancerous and healthy structures. Due to small size of the dataset and small ratio of tumor to healthy patch-instances, this step was necessary to ensure cancerous structures were captured in the CNN. We experimented with $K=100$ and $\alpha=0.5$ in reported experiments.

The VAE outlined in Table \ref{tab:experiments} was trained in an unsupervised manner using 135,000 patches extracted from the training set, in a similar manner to the CNN patch extraction method described above. We used Adam to optimize the VAE and stochastic gradient descent with a exponential decay function for InceptionNet \cite{Szegedy2016}. Both optimizers had an initial learning rate of $0.001$. All models were trained for $25$ epochs which took, on average, 2 days on two Nvidia Titan Xp GPUs. 

\subsubsection{Results}

\begin{figure}
\centering
    \begin{tabular}{cc}
	    \includegraphics[width=0.47\textwidth]{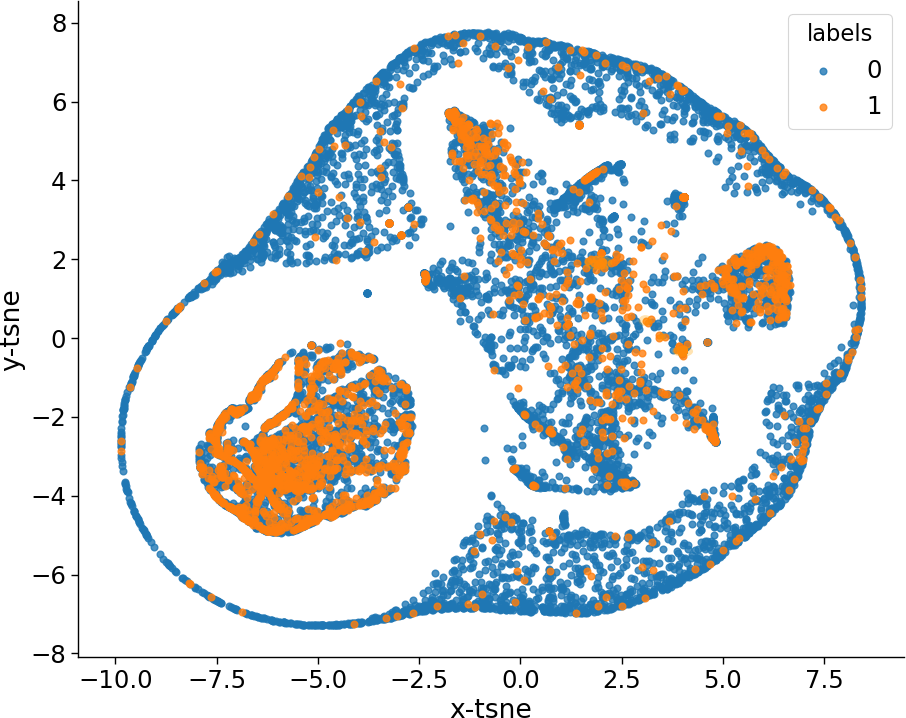} & 
	    \includegraphics[width=0.5\textwidth]{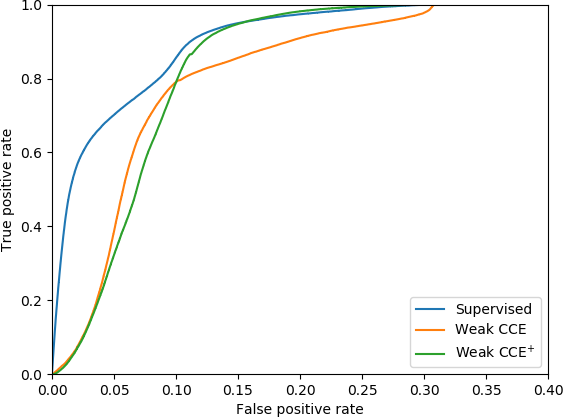}    
    \end{tabular}
	\caption{A t-SNE plot of the latent representation in the VAE for the Camelyon dataset (left) and ROC curves comparing an InceptionNet trained using three different configurations (right).}
	\label{fig:cam_results}
\end{figure}

A visualization of the latent representation in our trained VAE as a t-SNE plot is shown in Figure \ref{fig:cam_results} (left). Training patch-instances containing metastases are projected into this space and shown in orange and healthy patch-instances in blue. The t-SNE plot shows some clear clusters differentiating between the two classes indicating the VAE was able to capture features specific to each class without the need for expert annotations.  

ROC curves comparing a fully supervised approach (with accuracy labels per training instances), categorical cross-entropy (CCE) with weak labels and the proposed method, CCE$^{+}$,  is shown in Figure \ref{fig:cam_results} (right). Whilst CCE performed slightly better at lower sensitivities, our method was superior overall, reaching accuracy rates matching a fully supervised approach at higher sensitivities. This shows great promise for using weak labels for analysing large histology images.

\section{Conclusion}

In this paper, we described a novel weakly-supervised technique inspired by MIL for learning and leveraging a deep latent representation during training of a CNN. We used cluster-classes in a novel loss function to delineate between patches using weakly labeled bags. Our results show that our adapted loss function, CCE$^{+}$, can overcome issues with traditional CCE, particularly when large weakly bags are used. We also showed that this method can be used in digital pathology to analyse extremely high resolutional pathology images, and could potentially be used to automatically generate annotations in medicine where costs of experts annotators is extremely high.

\subsubsection*{Acknowledgments}

This research is funded by the Canadian Cancer Society (grant $319289$) and the National Cancer Institute of the National Institutes of Health (grant U24CA199374-01), and was enabled in part by support provided by ComputeCanada.

\bibliographystyle{plain}
\bibliography{nips}

\end{document}